\documentclass[runningheads]{llncs}
\usepackage[T1]{fontenc}
\usepackage{graphicx,verbatim}
\usepackage{amsmath}
\usepackage{enumitem}
\usepackage{amsfonts}
\usepackage{amssymb}
\usepackage{booktabs}
\usepackage{svg}
\usepackage{subcaption}
\usepackage[table]{xcolor}
\usepackage{booktabs}
\usepackage{multirow}
\usepackage{xurl}
\usepackage[hidelinks]{hyperref}
\newcommand{\cmark}{\checkmark}
\newcommand{\xmark}{\(\times\)}
\begin{document}

\title{A Principled Approach to Unsupervised Anomaly Detection}

\author{James Myles\inst{1} \and
Matthew Baugh\inst{1} \and
Johanna P. Müller\inst{2} \and
Bernhard Kainz\inst{1,2} \and
Yingzhen Li\inst{1,3}}

\authorrunning{J. Myles et al.}

\institute{
Imperial College London, UK \\ \email{james.myles24@imperial.ac.uk} \and
Friedrich-Alexander University Erlangen-Nürnberg, Germany \and
Nanyang Technological University, Singapore\
}

\maketitle

\begin{abstract}
Traditional unsupervised anomaly detection (UAD) methods are designed to flag or localise deviations from a normative distribution, ignoring the underlying generative mechanisms of the anomalies. Yet the nature of an anomaly is often as important as its presence. We reformulate UAD as a Bayesian inverse problem, in which the objective is to infer the most probable corruption responsible for each observation. Our framework yields a probabilistic anomaly score as the energy of the inferred corruption parameters, and serves as a principled recipe for developing new UAD algorithms. We derive several existing methods as instances of the general framework, each corresponding to the same energy score under different modelling choices. Experimentally, we study the framework's components in a controlled setting, and improve object-class AUROC on the MVTec AD dataset by $2.3$\% by adapting the underlying corruption model. Finally, we validate the framework on a brain MRI benchmark, achieving strong detection performance while producing estimates of pathology intensity, bias, and geometry. Code is available at \url{https://github.com/jgmyles/inverse-uad}.
\end{abstract}

\section{Introduction}
\label{sec: intro}
Anomaly detection is the task of identifying deviations from a normative (healthy) distribution. Unlike out-of-distribution detection, where anomalous samples bear little resemblance to the normative distribution, anomalies typically manifest as subtle perturbations of healthy samples~\cite{ood,UPD_study}. Nevertheless, most existing methods aim to identify or localise anomalies~\cite{glass,padim,ano_ddpm,rd} rather than reason about the underlying corruption. In contrast, characterising the observed corruption may offer a more interpretable diagnostic, which is particularly relevant in safety-critical domains such as healthcare~\cite{interpretability}.

In this paper, we formalise unsupervised anomaly detection (UAD) as a Bayesian inverse problem~\cite{bayesian_inverse_problems}, where the goal is to identify the form of corruption present in the data. We define a generative process that applies deterministic, parametrised corruptions to the healthy distribution, creating an explicit relationship between healthy and anomalous samples. This generative process induces an anomaly score given by the energy of the resulting joint distribution, which decomposes into three interpretable energy terms, each targeting distinct classes of anomalies. Beyond scoring, the proposed framework infers latent parametrisations of the corruption present in each observation. Practically, the framework provides a foundation to develop new UAD algorithms by selecting appropriate normative models, corruption families and priors that incorporate domain-specific prior knowledge.

\noindent\textbf{Contribution:} Building on this inverse problem formulation, we establish that prominent UAD methods are instances of our framework, differing only in three aspects: (i) the method of modelling the healthy distribution, (ii) the family of corruptions assumed to generate anomalies, and (iii) the prior they place on corruption parameters. Despite their differences, these methods ultimately employ the same probabilistic quantity to score anomalies. Experimentally, we study the framework's inference and scoring components in a controlled setting. On MVTec AD~\cite{mvtec}, we extend PaDiM~\cite{padim} to new corruption families, improving object-class AUROC by $2.3$\%. Finally, we verify the practical utility of our framework on a brain MRI benchmark~\cite{UPD_study}, inferring pathology intensity, bias and geometry while achieving competitive detection with the state-of-the-art.

\begin{figure*}[t]
\captionsetup[subfigure]{font=footnotesize, skip=2pt, labelformat=simple, labelsep=period, labelfont=bf}
\renewcommand\thesubfigure{\alph{subfigure}}
\centering
\begin{subfigure}[t]{0.151\textwidth}
\centering
\includegraphics[width=\linewidth]{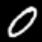}\\
\includegraphics[width=\linewidth]{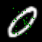}\\
\includegraphics[width=\linewidth]{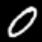}
\caption{Healthy}
\end{subfigure}
\hspace{0.003\textwidth}
\begin{subfigure}[t]{0.151\textwidth}
\centering
\includegraphics[width=\linewidth]{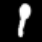}\\
\includegraphics[width=\linewidth]{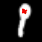}\\
\includegraphics[width=\linewidth]{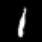}
\caption{Swollen}
\end{subfigure}
\hspace{0.003\textwidth}
\begin{subfigure}[t]{0.151\textwidth}
\centering
\includegraphics[width=\linewidth]{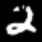}\\
\includegraphics[width=\linewidth]{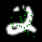}\\
\includegraphics[width=\linewidth]{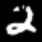}
\caption{Fractured}
\end{subfigure}
\hspace{0.003\textwidth}
\begin{subfigure}[t]{0.151\textwidth}
\centering
\includegraphics[width=\linewidth]{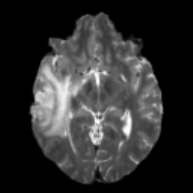}\\
\includegraphics[width=\linewidth]{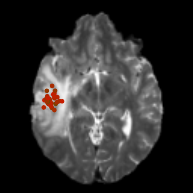}\\
\includegraphics[width=\linewidth]{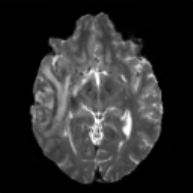}
\caption{BraTS-T2}
\end{subfigure}
\hspace{0.003\textwidth}
\begin{subfigure}[t]{0.151\textwidth}
\centering
\includegraphics[width=\linewidth]{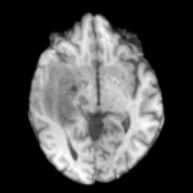}\\
\includegraphics[width=\linewidth]{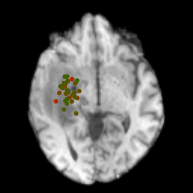}\\
\includegraphics[width=\linewidth]{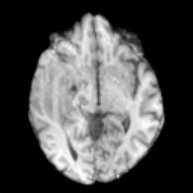}
\caption{BraTS-T1}
\end{subfigure}
\hspace{0.003\textwidth}
\begin{subfigure}[t]{0.151\textwidth}
\centering
\includegraphics[width=\linewidth]{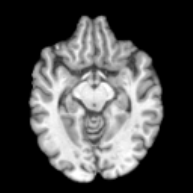}\\
\includegraphics[width=\linewidth]{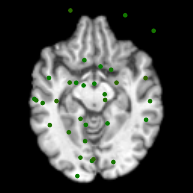}\\
\includegraphics[width=\linewidth]{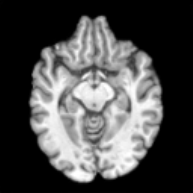}
\caption{ATLAS}
\end{subfigure}

\caption{
Qualitative corruption inference for MorphoMNIST~\cite{morpho} digits and BraTS-T2, BraTS-T1~\cite{brats1,brats2,brats3} and ATLAS~\cite{atlas} brain MRI scans. Rows show, from top to bottom: the observed image, anomaly parameter samples from the posterior estimator, and the de-corrupted sample corresponding to the posterior mode $x^*$. Posterior sample are indicated by dots coloured from green to red by inferred severity, defined as swelling strength $\gamma$ for MorphoMNIST and $\max(\alpha_{\mathrm{text}}, \left| \frac{\alpha_{\mathrm{bias}}}{b_{\max}} \right| )$ for brain MRI. The fractured digit lies outside the scope of the corruption function and is correctly left unaltered.
}
\label{fig:morphomnist_qualitative}
\end{figure*}

\noindent\textbf{Related Work.}
UAD methods span several paradigms. Reconstruction-based methods use deep model reconstruction error as an anomaly score~\cite{brain_cnn,vq_vae}. Feature-modelling methods detect anomalies in pre-trained feature spaces, either by modelling pixel-level features~\cite{padim,cflow_ad}, measuring feature-space reconstruction error~\cite{fae}, or comparing student and teacher networks~\cite{rd}. We show that some of these methods fall under our inverse problem framework. 
Closest to our work are methods that inject synthetic anomalies into healthy samples and use the result as a self-supervised signal, for example to reverse corruptions~\cite{cond_ddpm,disyre_v2,draem}, predict anomaly maps~\cite{many_tasks,glass,pii}, or learn representations separating healthy from corrupted samples~\cite{cut_paste}. In contrast, we model corruptions directly as a generative process and infer the latent corruption present in each observation. 
Inverse problems infer inputs from observations by reversing a forward mapping~\cite{inverse_problems}. They are often ill-posed, with no solution or multiple solutions. The Bayesian approach addresses this by defining probability distributions over variables, yielding a posterior distribution that encodes uncertainty~\cite{bayesian_inverse_problems}. Although many UAD methods use synthetic corruptions~\cite{disyre_v2,draem,many_tasks,glass,pii,cut_paste}, prior work has not cast this as a Bayesian inverse problem over corruption parameters.

\section{Anomaly Detection as a Bayesian Inverse Problem}

\label{sec: framework}
The central challenge in UAD is to detect subtle abnormalities in otherwise healthy samples. We model these irregularities as deterministic, parametrised corruptions to the healthy distribution $p_h(\cdot)$:
\begin{equation}
\label{eq: gen_process}
h \sim p_h(\cdot), \qquad x \sim p_x(\cdot), \qquad y = f(h, x), 
\end{equation}
where $f$ is a \textit{corruption function} and $x\in\mathcal{X}$ is a set of \textit{anomaly parameters} that describes the form of corruption applied to $h$. We propose to frame UAD as a Bayesian inverse problem~\cite{bayesian_inverse_problems}; given a sample $y$, the goal is to infer the most plausible explanation:
\begin{equation}
x^* = \text{argmax}_{x\in\mathcal{X}} \log p(x \mid y),
\end{equation}
offering a latent description of how $y$ deviates from the healthy distribution. Moreover, if $f$ is invertible, $x^*$ implies a healthy restoration $\hat{h} = f^{-1}(y, x^*)$. In Sect.~\ref{sec: morpho_mnist}, we explore three methods to obtain $x^*$: energy optimisation, regression and posterior approximation. \\
\noindent\textbf{Energy Scoring.}
Assuming $f$ is a diffeomorphism for any $x\in\mathcal{X}$, the product rule and change of variables theorem give the joint energy as
\begin{align}
\label{eq: post}
E(x, y) &:= -\log p(x, y) = -\log p(y|x) - \log p_x(x) \\
&
\label{eq: energy}
= \underbrace{-\log p_h(h)}_{E_{\text{healthy}}(h)} + \underbrace{\log \left| \det J_f(h, x) \right|}_{E_{\text{volume}}(h, x)} + \underbrace{-\log p_x(x)}_{E_{\text{anomaly}}(x)},
\end{align}
where $h = f^{-1}(y, x)$ is the healthy state corresponding to $x$ and $J_f$ denotes the Jacobian of $f$. Having inferred $x^*$, the corresponding minimised energy $E^*(y) = E(x^*, y)$ is an intuitive anomaly score; the anomaly energy $E_{\text{anomaly}}$ penalises large corrections, the healthy energy $E_{\text{healthy}}$ penalises corrected samples that remain far from the healthy manifold, and the volume energy $E_{\text{volume}}$ accounts for geometric distortion induced by the corruption map. For the identity corruption function and uniform anomaly prior, this reduces to the negative log-likelihood under the healthy distribution, $E^*(y) = -\log p_h(y) + C$, recovering likelihood-based approaches. In the following section, we show that prominent feature modelling methods can also be understood through the lens of our framework.

\noindent\textbf{Feature Modelling Methods as Inverse Problems.}
\label{sec: feature_modeling}
We show that CFLOW-AD~\cite{cflow_ad} and PaDiM~\cite{padim} arise from the proposed inverse problem framework, both scoring anomalies with the minimised energy $E^*(y)$ under specific corruption processes. We define the three components of the generative process in Eq.~\ref{eq: gen_process} that recover their anomaly scoring mechanisms. The healthy distribution is modelled using a normalising flow with invertible flow transformation $g$:
\begin{equation}
\label{eq: flow}
\log p_h(h) = \log p_z(g^{-1}(h)) + \log \left| \det J_{g^{-1}}(h) \right|,
\end{equation}
where $p_z = \mathcal{N}(\mu, \Sigma)$ is a Gaussian base distribution and $J_{g^{-1}}$ is the Jacobian of $g^{-1}$. We model anomalies as additive perturbations $x$ in the latent space of the flow:
\begin{equation}
f(h, x) = g(g^{-1}(h) + x),
\end{equation}
with isotropic Gaussian prior $p_x = \mathcal{N}(0, \epsilon I)$. For an observation $y$, let $w=g^{-1}(y)$ denote its latent representation. Under candidate corruption $x$, the corresponding healthy state is $h = g(w - x)$. Substituting this in Eq.~\ref{eq: energy}, applying the chain rule to the healthy energy and the inverse function theorem to the volume energy, the energy becomes
\begin{equation}
E(x, y) = - \log p_z(w-x) - \log p_x(x) - \log \left| \det J_{g^{-1}} (y)  \right|.
\end{equation}
The full derivation is provided in Lemma~\ref{lemma: cflow_energy} of Appendix~\ref{sec: cflow_deriv}. Since the volume term is independent of $x$, minimising the energy reduces to maximising $\log p_z(w-x) + \log p_x(x)$, which by Lemma~\ref{lemma: gaussian} of Appendix~\ref{sec: gauss_lemma} equals $\log \mathcal{N}(w; \mu, \Sigma + \epsilon I)$ up to an additive constant:
\begin{align}
E^*(y) &= -\max_x \left[\log p_z(w - x) + \log p_x(x) \right] - \log \left| \det J_{g^{-1}} (y)  \right| \\
\label{eq: flow_energy}
&= -\log \mathcal{N}(w; \mu, \Sigma + \epsilon I)  - \log \left| \det J_{g^{-1}} (y) \right| + C.
\end{align}
The regularised Mahalanobis distance score $M(y)$ used by PaDiM is obtained by taking the flow transformation to be the identity, corresponding to a Gaussian healthy distribution. In this case, the minimised energy given by
\begin{equation}
\label{eq: padim_energy}
E^*(y) = - \log \mathcal{N}(y; \mu, \Sigma + \epsilon I) + C = \frac{1}{2} M(y)^2 + C.
\end{equation}
The CFLOW-AD anomaly score corresponds to taking $\epsilon \rightarrow 0$ in Eq.~\ref{eq: flow_energy}:
\begin{equation}
E^*(y) \rightarrow - \log p_z(g^{-1}(y))  - \log \left| \det J_{g^{-1}} (y) \right| + C = -\log p_h(y) + C.
\end{equation}
The key insight is that both methods can be interpreted as inferring the most likely additive Gaussian corruption that explains an observation. CFLOW-AD performs this inference in the latent space of a learned normalising flow, whereas PaDiM performs the same inference directly in feature space.

\section{Experiments}
We evaluate our framework across controlled ablation studies on MorphoMNIST, corruption model generalisation on MVTec AD~\cite{mvtec}, and pathology detection on a brain MRI benchmark~\cite{UPD_study}. Detection performance is evaluated by Average Precision (AP, \%) for brain MRI and the Area Under the Receiver Operating Characteristic Curve (AUROC, \%) otherwise. To assess localisation, we report the best Sørensen–Dice index (Dice, \%) achieved over a threshold sweep. Unless otherwise stated, metrics are reported as mean $\pm$ standard deviation over $10$ random seeds and p-values are computed via a one-sided Wilcoxon signed-rank test with Holm correction for multiple tests.

\subsection{Parameter Recovery and Energy Scoring}
\label{sec: morpho_mnist}
In a controlled setting, we employ MorphoMNIST's \textit{Local} dataset~\cite{morpho}, which contains a mixture of healthy, fractured and swollen MNIST digits~\cite{mnist}. Digits are swollen by warping co-ordinates away from centre $r_0$ within a radius $R$ with strength $\gamma > 1$:
\begin{equation}
\varphi(r) = r_0 + (r - r_0) \left(\frac{\Vert r - r_0 \Vert}{R}\right)^{\gamma -1},
\end{equation}
obtaining the swollen image by bilinear resampling. Fractured digits are created by inserting three fractures at random locations along the digit normal. We use the swelling transformation as our corruption function with anomaly parametrisation $x = (c_x, c_y, \gamma) \in \left[0, 1 \right]^2 \times \mathbb{R}_{>1}$, where $c_x$, $c_y$ denote the centre co-ordinates scaled to the unit interval. We implement this transformation as a matrix multiplication $f(h, x) = A(x) h$, where $A(x)$ contains resampling weights. We intentionally omit fractures from the corruption function - we expect the healthy energy to detect these anomalies as no choice of swelling parameters $x$ can restore them to the healthy manifold. Following PaDiM, we model the healthy likelihood with Gaussians in feature space, extracting patches from a DINOv2~\cite{dinov2} backbone and randomly reducing to 256 dimensions~\cite{padim}. We place an exponential prior with rate $1$ over swelling strength, and a uniform prior over corruption centre. The energy terms are thus:
\begin{align}
E_{\text{healthy}}(h)
&= \sum_{i=1}^P \frac{1}{2}(h_i - \mu_i)^T \left(\epsilon I + \Sigma_i\right)^{-1} (h_i - \mu_i) + C_i,
\\
E_{\text{volume}}(h, x)
&= \log \left| \det A(x) \right|,
\\
E_{\text{anomaly}}(x)
&= \gamma + C,
\end{align}
where $\mu_i$ and $\Sigma_i$ are the healthy Gaussian components of patch $i$ estimated from healthy data, and $C$, $C_i$ are normalisation constants. For numerical stability, we use a regularised approximation to the volume energy:
\begin{equation}
E_{\text{volume}}(h, x) \approx \frac{1}{2} \log \left| \det A(x)^T A(x) + \epsilon I \right|.
\end{equation}
\noindent\textbf{Parameter Recovery.}
Deep generative models have been shown to outperform traditional Bayesian inverse approaches on inverse design problems~\cite{gen_inverse}. We follow this approach, training a CNN posterior estimator $q_\phi(x \mid y)$ using synthesised swellings to predict swelling strengths and centres. We regenerate the \textit{Local} dataset, retaining swelling parameters, and compare against a CNN regression baseline and gradient descent optimisation of the energy objective (Eq.~\ref{eq: energy}), measuring AUROC using strength as an anomaly score and Euclidean distance between inferred and ground-truth centres (pixels). Posterior approximation outperformed both methods on both metrics ($p<0.01$), achieving $100.0_{\pm0.0}$ AUROC and $0.87_{\pm0.06}$ pixel error, compared to $99.0_{\pm0.0}$/$82.0_{\pm1.8}$ AUROC and $1.90_{\pm0.14}$/$5.18_{\pm0.14}$ pixel error for regression and gradient descent respectively. The degraded performance of gradient descent indicates that posterior estimation offers a genuine advantage over direct optimisation of the energy objective for anomaly parameter inference. Qualitative inference examples are displayed in Fig.~\ref{fig:morphomnist_qualitative}.
\noindent\textbf{Energy Score Ablation Study.}
The second component of the framework is computing the energy score using the inferred anomaly parameters. We quantify the contribution of the three sub-scores by removing each component in turn. Intuitively, we expect the volume and anomaly energy terms to detect swollen samples based on the intensity of their inferred parameters. In contrast, the healthy energy should detect fractures, since the corruption model cannot restore them to the healthy distribution. Tab.~\ref{tab: energy_ablation} verifies this intuition; omitting the healthy energy degraded fracture detection ($p<0.01$), reducing AUROC by $31.3$, while dropping volume and anomaly energy terms hindered swollen digit detection ($p<0.01$). The full energy equation significantly outperformed all reported variants on the \textit{Local} dataset $(p<0.01)$, highlighting the interplay between the three terms. Notably, fracture detection for the energy score was on par with PaDiM, demonstrating that an exhaustive corruption function is not necessary due to the healthy energy term.

\subsection{Corruption Functions and Anomaly Priors}
\label{sec: mvtec}
\begin{table}[htbp]
\centering

\begin{minipage}[t]{0.45\textwidth}
    \centering
    \caption{Ablation on the three energy terms regarding AUROC on the \textit{Local} dataset and the swollen and fractured subsets. Energy scores were calculated using the same set of anomaly parameters inferred by a posterior estimator. PaDiM~\cite{padim} is provided as a baseline. Best scores are bold, second best are underlined.}
    \label{tab: energy_ablation}
    
    \vspace{0.5em}
    
    \resizebox{\linewidth}{!}{%
    \begin{tabular}{ccc ccc}
    \toprule
    \multicolumn{3}{c}{Energy Terms}
    &
    \multicolumn{3}{c}{AUROC (\%)} \\
    \cmidrule(lr){1-3}
    \cmidrule(lr){4-6}
    $E_{\mathrm{healthy}}$ &
    $E_{\mathrm{anomaly}}$ &
    $E_{\mathrm{volume}}$ &
    Swollen $\uparrow$ &
    Fractured $\uparrow$ &
    \textit{Local} $\uparrow$ \\
    \midrule
    \cmark & \cmark & \cmark & $\underline{99.3}_{\pm 0.1}$ & $92.3_{\pm 0.3}$ & $\mathbf{95.7}_{\pm 0.1}$ \\
    \xmark & \cmark & \cmark & $\mathbf{99.4}_{\pm 0.1}$ & $61.0_{\pm 2.4}$ & $79.9_{\pm 1.2}$ \\
    \cmark & \xmark & \cmark & $98.9_{\pm 0.1}$ & $\underline{92.4}_{\pm 0.3}$ & $\underline{95.6}_{\pm 0.1}$ \\
    \cmark & \xmark & \xmark & $94.8_{\pm 0.4}$ & $\mathbf{92.5}_{\pm 0.3}$ & $93.7_{\pm 0.2}$ \\
    \midrule
    \multicolumn{3}{c}{PaDiM~\cite{padim}} & $95.3_{\pm 0.2}$ & $92.0_{\pm 0.3}$ & $93.6_{\pm 0.2}$ \\
    \bottomrule
    \end{tabular}%
    }
\end{minipage}
\hfill
\begin{minipage}[t]{0.52\textwidth}
    \centering
    \caption{Image-level AUROC on MVTec AD~\cite{mvtec} object and texture classes for different choices of corruption function and anomaly prior. All combinations use the same Wide ResNet-50-2~\cite{wrn} backbone. Colours indicate improvement over PaDiM~\cite{padim}, which achieved $98.8_{\pm 0.2}$ and $92.9_{\pm 0.5}$ AUROC on texture and object classes respectively. $^*$ and $^{**}$ denote $p<0.05$ and $p<0.01$.}
    \label{tab: mvtec_heatmaps}
    
    \vspace{0.7em}
    
    \begin{minipage}[t]{0.48\linewidth}
        \centering
        \textbf{Texture}
        
        \vspace{0.3em}
        
        \resizebox{\linewidth}{!}{%
        \begin{tabular}{lcc}
        \toprule
        \multirow{2}{*}{\raggedright $f(h, x)$}
        & \multicolumn{2}{c}{$p_x$} \\
        \cmidrule(lr){2-3}
        & Gaussian & Laplace\\
        \midrule
        Add.
        & \cellcolor{blue!25.9}$98.9_{\pm0.2}^{*}$
        & \cellcolor{blue!0.0}$98.7_{\pm0.2}$ \\
        
        Mult.
        & \cellcolor{blue!39.1}$98.9_{\pm0.2}$
        & \cellcolor{blue!39.1}$98.9_{\pm0.2}$ \\
        
        Affine
        & \cellcolor{blue!55.0}$99.0_{\pm0.1}^{*}$
        & \cellcolor{blue!43.4}$99.0_{\pm0.2}$\\
        \bottomrule
        \end{tabular}%
        }
    \end{minipage}
    \hfill
    \begin{minipage}[t]{0.48\linewidth}
        \centering
        \textbf{Object}
        
        \vspace{0.1em}
        
        \resizebox{\linewidth}{!}{%
        \begin{tabular}{lcc}
        \toprule
        \multirow{2}{*}{\raggedright $f(h, x)$}
        & \multicolumn{2}{c}{$p_x$} \\
        \cmidrule(lr){2-3}
        & Gaussian & Laplace\\
        \midrule
        Add.
        & \cellcolor{blue!1.5}$93.1_{\pm0.5}^{**}$
        & \cellcolor{blue!0.0}$93.1_{\pm0.4}^{*}$ \\
        
        Mult.
        & \cellcolor{blue!41.5}$94.7_{\pm0.2}^{**}$
        & \cellcolor{blue!25.1}$94.1_{\pm0.3}^{**}$ \\
        
        Affine
        & \cellcolor{blue!47.6}$95.0_{\pm0.3}^{**}$
        & \cellcolor{blue!55.0}$95.2_{\pm0.2}^{**}$ \\
        \bottomrule
        \end{tabular}%
        }
    \end{minipage}
\end{minipage}

\end{table}
We illustrate how the proposed framework can be used to develop new UAD algorithms. In particular, we build on the theory from Sect.~\ref{sec: feature_modeling}, which demonstrated that PaDiM~\cite{padim} implicitly models anomalies as additive Gaussian noise. We modify this assumption, exploring two alternative corruption functions: multiplicative $f(h, x) = h \odot e^x$ and affine $f(h, x) = h \odot e^{x_a} + x_b$, as well as a Laplace prior whose variance is matched to that of the Gaussian prior ($\epsilon=0.01$). In all cases, we retain PaDiM's healthy distribution modelling, extracting features from a Wide ResNet-50-2~\cite{wrn} backbone and performing random reduction to 550 dimensions~\cite{padim}. The minimised energy score is used as a pixel-level anomaly score, obtained via 250 steps of gradient descent. The exception is the additive Gaussian case, where we use the closed-form solution in Eq.~\ref{eq: padim_energy}. The maximum pixel-level score is used as an image-level anomaly score. We evaluate these modifications on the MVTec AD~\cite{mvtec} dataset and report detection performance for object and texture classes in Tab.~\ref{tab: mvtec_heatmaps}. For object classes, all variants outperformed PaDiM ($p<0.05$), with multiplicative and affine corruption yielding the largest improvements ($p<0.01$). Introducing a Laplace prior alongside affine corruption achieved the strongest detection, improving PaDiM by $2.3$ AUROC. Texture classes were less sensitive to modelling choices; only additive and affine Gaussian corruption significantly outperformed PaDiM ($p<0.05$). These results highlight that modelling choices can have a material impact on detection performance. The proposed framework provides a mechanism for adapting these choices to the target domain.

\subsection{Brain MRI}
\label{sec: brain_mri}
We evaluate our framework on the brain MRI benchmark from~\cite{UPD_study}, closely adhering to their preprocessing pipeline. The benchmark utilises the following datasets:
\begin{itemize}
\item \textbf{CamCAN}: The Cambridge Centre for Ageing and Neuroscience (CamCAN) dataset~\cite{camcan}, containing 653 healthy subjects with paired \mbox{$T_1$}- and \mbox{$T_2$}-weighted scans.
\item \textbf{ATLAS}: The Anatomical Tracings of Lesions After Stroke (ATLAS) dataset~\cite{atlas}, containing 655 \mbox{$T_1$}-weighted volumes with stroke lesion annotations.
\item \textbf{BraTS}: The Brain Tumor Segmentation (BraTS) dataset~\cite{brats1,brats2,brats3}, containing 369 subjects with brain tumour annotations. We denote the \mbox{$T_1$}- and \mbox{$T_2$}-weighted subsets BraTS-T1 and BraTS-T2, respectively.
\end{itemize}
CamCAN is used for training, while the remaining datasets are used to assess detection performance. Inspired by~\cite{disyre_v2}, our corruption function inserts masked texture into slices $h$:
\begin{equation}
\label{eq: brain_corruption}
f(h, x) = h + m(x) \odot \left(\alpha_{\mathrm{text}} (t(x) - h) + \alpha_{\mathrm{bias}} \right),
\end{equation}
where $\alpha_{\mathrm{text}} \in \left[0.0, 1.0 \right]$, $\alpha_{\mathrm{bias}} \in \left[-b_{\max}, b_{\max} \right]$, $m(x)$ is a shape mask and $t(x)$ is a foreign texture patch. The shape boundary is defined in mask-centred polar co-ordinates; at angle $\theta$, the radius is given by a truncated Fourier series:
\begin{equation}
r_\Omega(\theta) = r_0 \exp \left\{ \frac{1}{K} \sum_{k=1}^K a_k \sin(k\theta) + b_k \cos(k \theta) \right\}.
\end{equation}
The shape mask corresponds to the interior of $\Omega$ with softening applied at the boundaries. Shape and texture patches are scaled horizontally by a factor $s$, resized to area $A$ and translated to centre $c$ before applying Eq.~\ref{eq: brain_corruption}. We denote this operator as $\mathcal{W}_{x}$.
\begin{figure}[t]
\includegraphics[width=\textwidth]{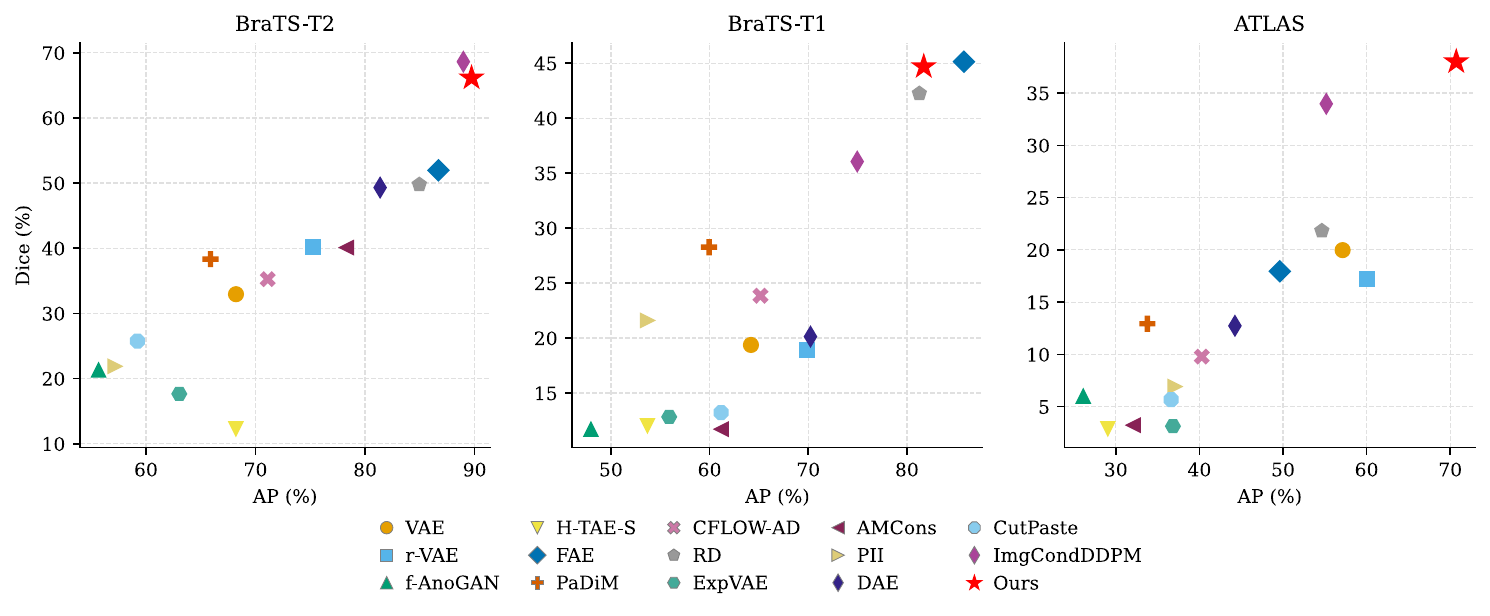}
\caption{AP (x-axis) against Dice (y-axis) on the brain MRI benchmark~\cite{UPD_study}. Baseline metrics are taken from~\cite{UPD_study} and~\cite{cond_ddpm}. We note that DISYREv2~\cite{disyre_v2} reported $73.0$, $51.0$ and $45.0$ Dice on BraTS-T2, BraTS-T1 and ATLAS respectively, but did not provide slice-level detection metrics. 'Ours' denotes our method, with results reported a single seed.
}
\label{fig: brain_results}
\end{figure}
We treat the texture patch as a nuisance anomaly parameter and train a CNN posterior density estimator $q_\phi(x \mid y)$ to infer geometry parameters: $s, A, c, \left\{a_k\right\}_{k=1}^K$ and $\left\{b_k\right\}_{k=1}^K$, intensity $\alpha_{\mathrm{text}}$ and bias $\alpha_{\mathrm{bias}}$. We also train a restoration model $g_\theta(y, x)$ to invert the corruption. The training loss is given by
\begin{equation}
    \mathcal{L} = \mathbb{E}_{h, x} \left[-\log q_{\phi}(x \mid y) + \lambda_h \lVert g_\theta(y, x) -h\rVert_2^2 \right], \qquad y = f(h, x).
\end{equation}
We construct anomaly maps by adding the restoration residual to the mean predicted shape, scaled by the mean predicted intensity $\bar{\alpha}_{\mathrm{text}}$ and bias $\bar{\alpha}_{\mathrm{bias}}$:
\begin{equation}
\label{eq: brain_anomaly_map}
M(y) = \left|y - g_\theta(y, x^*) \right| + \left[\bar{\alpha}_{\mathrm{text}} + \frac{\bar{\alpha}_{\mathrm{bias}}}{b_{\max}} \right] \frac{1}{N} \sum_{i=1}^N \mathcal{W}_{x_i} (m(x_i)),
\end{equation}
where $\left\{x_i\right\}_{i=1}^N \sim q_\phi(\cdot \mid y)$ and $x^*$ is the posterior mode. The anomaly score is the sum over anomaly map pixels. In practice, we take $b_{\max} =0.4$, $r_0=0.65$, $K=3$, $\lambda_h = 0.1$ and $N=128$. Localisation and detection metrics are presented in Fig.~\ref{fig: brain_results}. The proposed method achieved competitive results with state-of-the-art methods on all datasets, in particular achieving $38.0$ Dice on ATLAS, surpassing all methods except DISYREv2~\cite{disyre_v2}. Moreover, the proposed framework yields interpretable pathology parameters. The faithfulness of the inferred parameters is quantified in Tab.~\ref{tab:mri_param_recovery}. Across all datasets, inferred lesion size showed a moderate-to-strong correlation with ground-truth size, with the strongest correspondence on BraTS-T2 ($\rho=0.735$). Centre localisation was also most effective on BraTS-T2, with $71.7$\% of inferred centres lying with 4 pixels of the lesion mask. The weaker localisation on ATLAS ($29.4$\%) likely reflects the prevalence of small anomalies \cite{UPD_study}, for which accurate localisation is more challenging. Intensity bias showed weak-to-moderate correlation on ATLAS and BraTS-T2, but only weak correlation on BraTS-T1 ($\rho=0.120$), consistent with BraTS-T1 lesions often exhibiting limited intensity contrast relative to healthy tissue \cite{UPD_study}. Inference examples are displayed qualitatively in Fig.~\ref{fig:morphomnist_qualitative}.
\begin{table}[t]
\centering
\caption{Quantitative parameter recovery metrics on the brain MRI benchmark, computed over anomalous slices only. Size and bias correlation report Spearman rank correlation between inferred and ground-truth values, where ground-truth bias is the mean intensity difference between the lesion mask and the corresponding region in matched healthy slices. Centre localisation is the proportion of samples where the inferred anomaly centre falls within 4 pixels of the mask boundary. Confidence intervals are 95\% bootstrap intervals.}
\label{tab:mri_param_recovery}
\setlength{\tabcolsep}{6pt}
\begin{tabular}{lccc}
\toprule
Dataset & Size correlation $\uparrow$ & Bias correlation $\uparrow$ & Centre localisation (\%) $\uparrow$ \\
\midrule
ATLAS     & 0.600 [0.591, 0.608] & 0.323 [0.311, 0.336] & 29.4 \\
BraTS-T1  & 0.558 [0.549, 0.567] & 0.120 [0.107, 0.135] & 54.3 \\
BraTS-T2  & 0.735 [0.726, 0.743] & 0.492 [0.482, 0.503] & 71.7 \\
\bottomrule
\end{tabular}
\end{table}

\section{Limitations and Future Work}
Our posterior estimators $q_\phi(x \mid y)$ are trained on synthetic corruptions, so their reliability depends on how  accurately these corruptions resemble real anomalies. This risk is partly mitigated by the healthy energy, which - as shown in Tab.~\ref{tab: energy_ablation} - continues to flag anomalies (e.g. fractures) that fall outside the scope of the corruption function. More broadly, designing faithful corruption functions and anomaly priors may be practically challenging or infeasible. In such cases, our experiments indicate that simple corruption families, such as additive Gaussian noise in feature space, can be effective. We view our framework as a foundation for incorporating domain knowledge where possible, rather than a method that requires it universally. For brain MRI, we opt for a heuristic anomaly score (Eq.~\ref{eq: brain_anomaly_map}) rather than the minimised energy score, which would require accurate density modelling of complex anatomy. Although the energy score is not an essential component of the framework, exploring diffusion models or normalising flows is a natural avenue for future work. Finally, we do not explore sensitivity to prior specification; given the number of reported experiments, we leave systematic sensitivity analysis to future work.

\section{Conclusion}
We recast UAD as a Bayesian inverse problem that infers the corruption underlying each observation. This formulation recovers interpretable anomaly descriptions and motivates an anomaly score that decomposes into healthy, volume and anomaly energy terms. We unified popular anomaly detection methods, including PaDiM~\cite{padim} and CFLOW-AD~\cite{cflow_ad}, under our framework and illustrated how they can be generalised to new forms of corruption. Experiments on MorphoMNIST~\cite{morpho}, MVTec AD~\cite{mvtec}, and brain MRI show accurate parameter recovery, improved object-class detection, and a step towards characterising, not just detecting, pathology. We hope this perspective will provide a foundation for future work on interpretable and domain-specialised anomaly detection.

\begin{credits}

\subsubsection*{Data Availability Statement.}
All datasets used in this study are publicly available. MorphoMNIST is available at \url{https://github.com/dccastro/Morpho-MNIST}; MVTec AD at \url{https://kaggle.com/datasets/ipythonx/mvtec-ad}; CamCAN at \url{https://opendata.mrc-cbu.cam.ac.uk/projects/camcan/}; BraTS 2020 at \url{https://kaggle.com/datasets/awsaf49/brats20-dataset-training-validation}; and ATLAS R2.0 at \url{https://fcon_1000.projects.nitrc.org/indi/retro/atlas.html}. Access to individual datasets is subject to their respective licences and data-use agreements.

\subsubsection{\ackname}

Support was received from ERC projects MIA-NORMAL 101083647 and J.M. is supported by a UKRI DTP award.

\subsubsection{\discintname}
The authors have no relevant competing interests.
\end{credits}

\bibliographystyle{splncs04}
\bibliography{references}

\appendix

\section{Feature Modelling Derivations}

\subsection{Normalising Flow Energy Derivation}
\label{sec: cflow_deriv}
\begin{lemma}
\label{lemma: cflow_energy}
The energy of the normalising flow corruption process in Sect.~\ref{sec: feature_modeling} is given by:
\begin{equation}
E(x, y) = - \log p_z(w-x) - \log p_x(x) - \log \left| \det J_{g^{-1}} (y)  \right|.
\end{equation}
\noindent\textbf{Proof}: The inverse corruption function is given by $h = f^{-1}(y, x) = g(w -x)$ where $w=g^{-1}(y)$. The healthy energy is obtained by substituting $h$ into the normalising flow likelihood:
\begin{align}
E_{\text{healthy}}(h, x) &= - \log p_z(g^{-1}(h)) - \log \left| \det J_{g^{-1}} (h) \right| \\
&= - \log p_z(w - x) - \log \left| \det J_{g^{-1}}(g(w-x)) \right| \\
&= - \log p_z(w - x) + \log \left| \det J_g (w-x) \right|,
\end{align}
where we used the inverse function theorem in the last line. Using the alternate expression for the corruption Jacobian in terms of its inverse and applying the chain rule, the volume energy is
\begin{align}
E_{\text{volume}}(h, x) &= -\log \left| \det J_{f^{-1}} (y, x) \right| \\
&= -\log \left| \det J_g (w - x) \right| - \log \left| \det J_{g^{-1}}(y) \right|.
\end{align}
Finally, the anomaly energy is simply $E_{\text{anomaly}}(x) = -\log p_x(x)$. Summing the three terms yields the desired expression.
\end{lemma}

\subsection{Gaussian Maximisation Lemma}
\label{sec: gauss_lemma}
\begin{lemma}
\label{lemma: gaussian}
Given two Gaussian distributions
\begin{equation}
p_h(\cdot) = \mathcal{N}(\cdot; \mu, \Sigma_h), \qquad p_x(\cdot) = \mathcal{N}(\cdot; 0, \Sigma_x),
\end{equation}
the following identity holds:
\begin{equation}
\max_x \left[ \log p_h(y - x) + \log p_x(x) \right] = \log \mathcal{N}(y; \mu, \Sigma_h + \Sigma_x) + C.
\end{equation}
\noindent\textbf{Proof}: Expanding the Gaussian densities yields
\begin{align}
\log p_h(y - x) + \log p_x(x) &= -\frac{1}{2} \left[(y - x - \mu)^T \Sigma_h^{-1} (y - x - \mu) + x^T \Sigma_x^{-1} x \right] + C^\prime \\
&
\label{eq: gauss_sum}
= -\frac{1}{2} \underbrace{\left[(v - x)^T \Lambda_h (v - x) + x^T \Lambda_x x \right]}_{R(x)} + C^\prime,
\end{align}
where $\Lambda_h = \Sigma_h^{-1}$, $\Lambda_x = \Sigma_x^{-1}$, and $v = y - \mu$. Setting the derivative to zero:
\begin{equation}
\frac{dR}{dx} = -2\Lambda_h(v - x) + 2\Lambda_x x = 0 \implies x^* = (\Lambda_h + \Lambda_x)^{-1} \Lambda_h v.
\end{equation}
The Hessian $2(\Lambda_h + \Lambda_x) \succ 0$ confirms $x^*$ is a global minimum. At the optimum, stationarity gives $\Lambda_h(v - x^*) = \Lambda_x x^*$, so
\begin{align}
R(x^*) &= (v - x^*)^T \Lambda_h(v - x^*) + {x^*}^T \Lambda_x x^* \\
        &= (v - x^*)^T \Lambda_x x^* + {x^*}^T \Lambda_x x^* \\
        &= v^T \Lambda_x x^*.
\end{align}
Finally, substituting in the expression for $x^*$, we obtain
\begin{align}
R(x^*) &= v^T \Lambda_x (\Lambda_h + \Lambda_x)^{-1} \Lambda_h v \\
&= v^T (\Sigma_h + \Sigma_x)^{-1} v \\
&= (y - \mu)^T (\Sigma_h + \Sigma_x)^{-1} (y - \mu),
\end{align}
where the penultimate step uses $(A^{-1} + B^{-1})^{-1} = A(A+B)^{-1}B$. Substituting $R(x^*)$ into Equation \ref{eq: gauss_sum} completes the proof.
\end{lemma}

\end{document}